\documentclass[10pt,journal,compsoc]{IEEEtran}
\ifCLASSOPTIONcompsoc
  \usepackage[nocompress]{cite}
\else
  \usepackage{cite}
\fi
\ifCLASSINFOpdf
   \usepackage[pdftex]{graphicx}
\else
\fi
\usepackage{amsmath}
\usepackage{array}
\usepackage{url}
\usepackage{amssymb}
\usepackage{amsfonts}
\usepackage{multirow}
\usepackage{booktabs}
\def\ie{\emph{i.e.}}
\def\eg{\emph{e.g.}}

\def\etal{\textit{et al.~}}

\begin{document}
%
\title{Cross-Modality Brain Tumor Segmentation via Bidirectional Global-to-Local Unsupervised Domain Adaptation}
%
%
%
%

\author{Kelei~He,~
        Wen~Ji,~
        Tao~Zhou,~
        Zhuoyuan~Li,~
        Jing~Huo,~
        Xin~Zhang,~
        Yang~Gao,~\\
        Dinggang~Shen,~\IEEEmembership{Fellow,~IEEE},~
        Bing~Zhang,~
        and~Junfeng~Zhang

\IEEEcompsocitemizethanks{
\IEEEcompsocthanksitem K. He, Z. Li and J. Zhang are with Medical School of Nanjing University, Nanjing, China. K. He, W. Ji, Z. Li, Y. Gao and J. Zhang are also with the National Institute of Healthcare Data Science at Nanjing University, Nanjing, China. J. Wen, J. Huo and Y. Gao are with the State Key Laboratory for Novel Software Technology, Nanjing University, Nanjing, China. 
\IEEEcompsocthanksitem T. Zhou is with School of Computer Science and Technology, Nanjing University of Science and Technology, Nanjing, China. 
\IEEEcompsocthanksitem B. Zhang and X. Zhang are with Department of Radiology, Nanjing Drum Tower Hospital, Nanjing University Medical School, China.
\IEEEcompsocthanksitem D. Shen is with School of Biomedical Engineering, ShanghaiTech University, Shanghai, China. He is also with Department of Research and Development, Shanghai United Imaging Intelligence Co., Ltd., Shanghai, China. He is also with Department of Artificial Intelligence, Korea University, Seoul, Republic of Korea.
}
\thanks{* Corresponding authors: Junfeng Zhang (jfzhang@nju.edu.cn); Bing Zhang (zhangbing\_nanjing@nju.edu.cn)}}

\markboth{Journal of \LaTeX\ Class Files,~Vol.~14, No.~8, August~2015}%
{Shell \MakeLowercase{\textit{et al.}}: Bare Demo of IEEEtran.cls for Computer Society Journals}
%



\IEEEtitleabstractindextext{%
\begin{abstract}
Accurate segmentation of brain tumors from multi-modal Magnetic Resonance (MR) images is essential in brain tumor diagnosis and treatment. However, due to the existence of domain shifts among different modalities, the performance of networks decreases dramatically when training on one modality and performing on another, e.g., train on T1 image while perform on T2 image, which is often required in clinical applications. This also prohibits a network from being trained on labeled data and then transferred to unlabeled data from a different domain. To overcome this, unsupervised domain adaptation (UDA) methods provide effective solutions to alleviate the domain shift between labeled source data and unlabeled target data. In this paper, we propose a novel Bidirectional Global-to-Local (BiGL) adaptation framework under an UDA scheme. Specifically, A bidirectional image synthesis and segmentation module is proposed to segment the brain tumor using the intermediate data distributions generated for the two domains, which includes an image-to-image translator and a shared-weighted segmentation network. Further, a global-to-local consistency learning module is proposed to build robust representation alignments in an integrated way. Extensive experiments on a multi-modal brain MR benchmark dataset demonstrate that the proposed method outperforms several state-of-the-art unsupervised domain adaptation methods by a large margin, while a comprehensive ablation study validates the effectiveness of each key component. The implementation code of our method will be released at \url{https://github.com/KeleiHe/BiGL}.
\end{abstract}

}

\maketitle

\IEEEdisplaynontitleabstractindextext

%
\IEEEpeerreviewmaketitle

\IEEEraisesectionheading{\section{Introduction}\label{sec:introduction}}

%
%
%
%
\IEEEPARstart{S}egmentation of the tumors and surrounding infected area in multi-modal brain Magnetic Resonance (MR) images is essential for the diagnosis and treatment of various types of brain cancer, including glioma, meningioma, etc. \cite{Menze2010,havaei2017brain,zhou2020hi}.
Deep neural networks have demonstrated promising results in the automatic segmentation task~\cite{chen2016voxresnet,fu2019dual}. Typically, the brain tumor segmentation methods are trained in a fully-supervised and multi-modal manner, in which sufficient delineations are required as the guidance.
However, analyzing brain MR images often faces large domain shifts \cite{stacke2019closer,pooch2019can}, \eg, the data consists of multiple modalities, acquired from different hospitals and scanners, with varied manufacturers and scanning parameters. As a result, the performance of trained models will dramatically decrease in the following two situations: 1) applying the models trained from one modality to another; and 2) performing on data from different machines with different protocols.

\begin{figure}[!t]
  \centering
  \includegraphics[width=\linewidth]{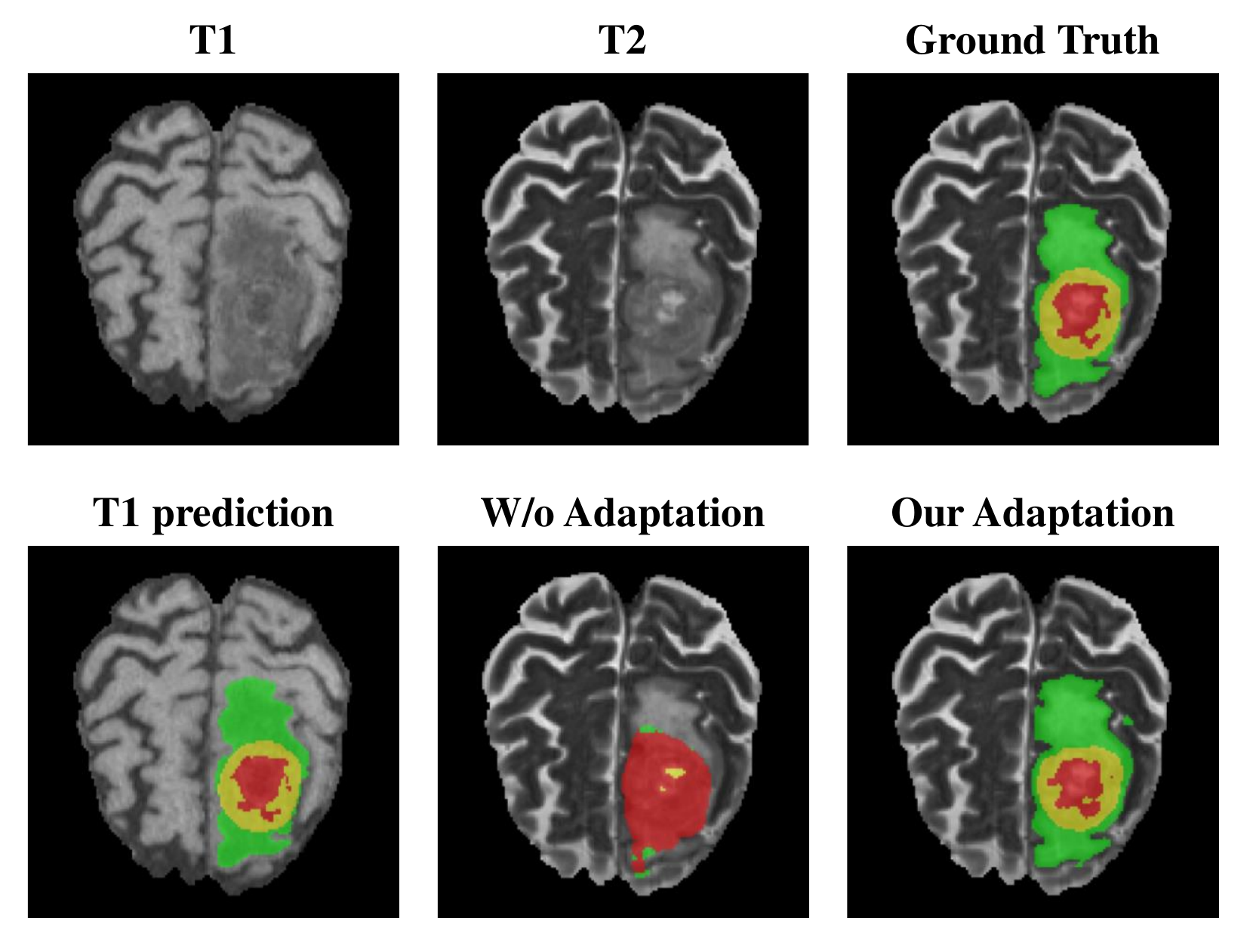}
    \caption{\label{fig:spotlight} The segmentation performance of brain tumors on a typical case. The models are trained using MR images of T1 sequence. The segmentation results on T2 images with the proposed BiGL framework are significantly better than without adaptation.}
\end{figure}

On the other hand, delineating the brain tumors in multi-modal images is unacceptable in clinical situation for two reasons: 1) it requires substantial time and clinical expertise, and 2) the tumors and infected regions are not clear or even absence in some modalities. Therefore, it would be highly desirable for trained networks to be able to adapt between modalities, with segmentations only labeled on one modality. This would also make the segmentation methods much more user friendly for clinicians, as it saves them potential time. However, this is a very challenging task due to the varied contexts in different modalities \cite{wilson2020survey}. 
To tackle this challenge, the unsupervised domain adaptation (UDA) methods \cite{cycada,output,SIFA-TMI,zhang2018task,moriya2018unsupervised} have been explored and developed. The UDA methods aim to improve the model generalization ability when learning from the labeled data (\ie, source domain) and performing on the different unlabeled data (\ie, target domain) with domain shifts. 
Although UDA has proven effective, existing methods still have several limitation. First, some of these methods partially leverage synthesized images to achieve domain adaptation \cite{cycada,SIFA-TMI}, \ie, they only use the synthesized target images to construct the model. However, the synthesized source images can also take effect on this kind of problem. Second, they align inefficient representations for the adaptation. The structured output assumption in \cite{output} is not applicable to some tasks, \eg, medical image segmentation. Besides, most previous methods \cite{cycada,SIFA-TMI} use the redundant features for domain alignment, which inhibits the performance of UDA.
Third, most methods were originally designed for the task of organ segmentation, which is easier than tumor segmentation, as tumors are not of structural appearances. We report the performance of these networks as well as a baseline network without adaptation in experiments (as reported in Section \ref{Sec:Exp}), and show they are unable to generate high-quality segmentation.

To overcome the aforementioned limitations, we propose a framework that can accurately delineate brain tumors in one sequence of MR images (\ie, the target domain) when trained on another sequence of MR images (\ie, the source domain), and vise versa. In the training stage, the proposed method randomly selects two unpaired images (one image from each domain) as the input. Then, the images are send to the framework which consists of two main components. Firstly, to reduce the gap between the two domains, a bidirectional image-to-image generation and segmentation module is introduced. It first generates one synthesized image for each original image, and then segments all the images (\ie, the original and synthesized images) using a encoder-decoder architecture. The features and outputs of the images are further aligned by the novel global-to-local adaptor (GTA) applied to the network. From a global perspective, the adaptor aligns the real and synthesized images of one subject in the output space, by assuming that they share the same label. It also aligns the real and synthesized images of different subjects through the commonly used feature alignment. However, tumors are often located in small arbitrary regions of the image, making tumor segmentation a more difficult task than organ segmentation. Therefore, in the GTA, we propose a local alignment by utilizing the attention maps captured from the features. This attention-based alignment helps transfer the most discriminate part of features through domains, by assuming the silences of the images are consistent inter-domain, thus enhancing the adaptation ability. Moreover, inspired by the work in \cite{cycada,output}, we conduct adversarial learning to build distribution consistencies between the aligned representations. 

To conclude, the contributions of our proposed method are three-fold:
\begin{itemize}
    \item We propose a novel bidirectional global-to-local (BiGL) unsupervised domain adaptation method to tackle the challenging problem of cross-modal brain tumor segmentation, in which unsupervised domain adaptation, adversarial learning, attention mechanism and brain tumor segmentation are integrated into a unified framework.  
 
    \item The global-to-local adaptation module is proposed to reduce domain gap between the source and target domains as well as between the synthesized and real images, which can effectively improve the generalization ability of UDA and thus boost the segmentation performance.
    
    \item Experimental results on a brain tumor benchmark dataset (\ie, BraTS' 19 \cite{bakas2018identifying}) demonstrate the effectiveness of our BiGL method, which significantly outperforms other state-of-the-art UDA methods. And experiments on another benchmark dataset (\ie, MMWHS \cite{zhuang2016multi}) for cross-modality (\ie, CT to MR) whole heart segmentation shown the generalization ability of the proposed method.
\end{itemize}

The following sections are organized as follows: Section \ref{sec:related} reviews the related work of the proposed method; Section \ref{sec:method} introduces the detailed architecture of the proposed BiGL method for cross-modal brain tumor segmentation; Section \ref{Sec:Exp} reports the experiments of the proposed method on a brain tumor segmentation dataset and a benchmark heart structural segmentation dataset; Section \ref{sec:conclusion} concludes this work and gives possible future directions.

\section{Related Work}
\label{sec:related}

We briefly review two types of works that are most related to our method, including 1) deep learning-based methods for brain tumor segmentation, and 2) unsupervised domain adaptation methods and its applications in medical images.

\subsection{Deep Learning-based Brain Tumor Segmentation}

Deep learning methods have achieved remarkable results in the application of brain tumor segmentation. We roughly divide the related segmentation methods into two categories. 

1) Single-prediction methods that predict the label of a single pixel over an input image patch. For example, Pereira \etal \cite{pereira2016brain} propose a automatic brain tumor segmentation method based on a specifically designed CNN architecture with data augmentation techniques, \eg, normalization and rotation. Havaei \etal \cite{HAVAEI201718} proposed to tackle the brain tumor segmentation task via cascaded CNNs. Specifically, they solved the label imbalance problem using a two-stage framework, and utilized a two-pathway CNN architecture to further leverage multi-scale features. As CNN-based methods are quite time consuming with redundant operations, the FCN-based method to densely predict segmentation is more popular in recent years.

2) Densely-prediction methods use a fully-convolutional network (FCN). For example, Dong \etal \cite{dong2017automatic} uses the popular FCN architecture in medical image segmentation, \ie, U-Net, to segment brain tumors in MR images.
By leveraging the U-Net and conditional random field method, Chen \etal \cite{CHEN201990} proposed a two-pathway network to incorporate multi-scale inputs for tumor segmentation.
These methods show the effectiveness of the U-Net under fully-supervision, especially with multi-modality data annotations. However, the segmentation performance under the weakly-supervised setting will be quite poor. In this case, the unsupervised domain adaptation applied to solve the brain tumor segmentation problem is required.

\subsection{Unsupervised Domain Adaptation Methods}

The unsupervised domain adaptation methods can be roughly categorized into two types, \ie, the one-step UDA methods and the multi-step domain adaptation methods. The one-step UDA methods often simply align the feature distributions of the two domains by minimizing their distances \cite{ganin2016domain,yang2019unsupervised}.
For example, Tsai \etal \cite{output} proposed a multi-level adversarial architecture to adapt structured output from a source to target domain for semantic segmentation. This method performs alignment on the original domain data, but it is difficult for the network to minimize the huge domain discrepancy. 

By contrast, the multi-step UDA methods often introduce intermediate domains between the source and target domain in order to get a more smooth domain alignment. In existing methods, these intermediate domains are often generated by generative adversarial learning methods. For instance, Hoffman \etal \cite{cycada} proposed to use an unsupervised image translator (\ie, Cycle-GAN \cite{zhu2017unpaired}) to generate fake target images that have closer feature distributions to the target domain. Then, the unpaired fake image and target image were utilized to build a framework that can effectively guide the adaptation of the learned network.

A few attempts have also tried to address the UDA problem in medical image applications. \cite{zhang2018task,chen2020deep}. For instance, \cite{SIFA} proposed a UDA framework that uses cycle-consistencies to generate a faked source image, and then performs alignment on the fake and target image pair in the output space. The work combines the advantages of the existing methods proposed in \cite{cycada} and \cite{output}. However, these methods used partially the generated data information, which lead to limited domain alignment performance.

\section{Methods}
\label{sec:method}

\subsection{Overview of Framework}

\begin{figure*}[!t]
  \centering
  \includegraphics[width=\linewidth]{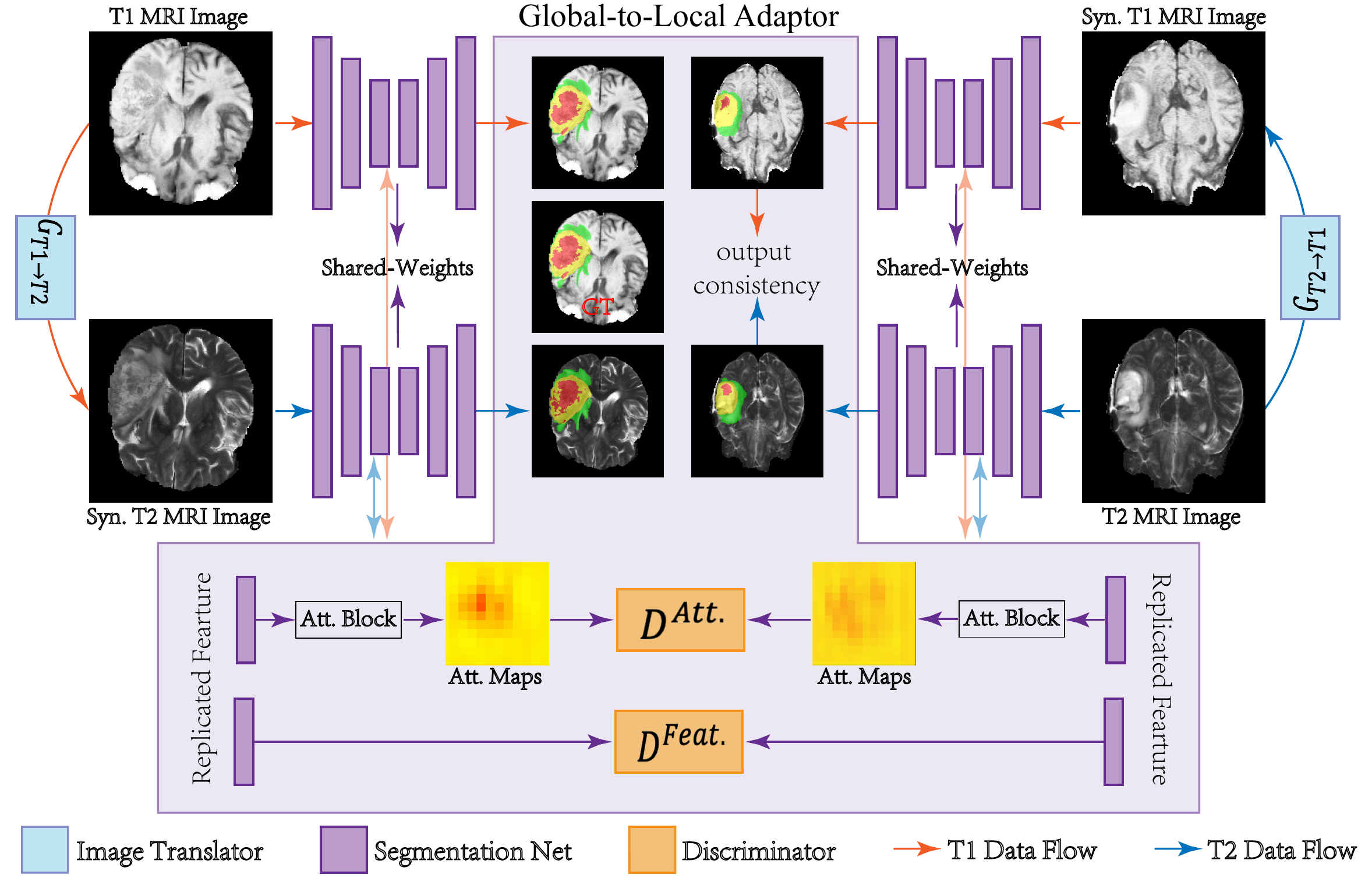}
    \caption{\label{fig:framework} The architecture of our proposed BiGL framework. We use the task of $T_1 \to T_2$ as an example, which regards the annotated $T_1$ MR images as the source domain, and non-annotated $T_2$ images as the target domain. $G_{T_* \to T_*}$ denote the generator for cross-domain image synthesis. "GT" denote the ground-truth.
    }
\end{figure*}

For simplicity, we construct the proposed BiGL framework between $T_1$ (as source domain) and $T_2$ (as target domain) MR images as an example to illustrate the overview architecture, as shown in Fig. \ref{fig:framework}. The proposed framework uses two unpaired images as the inputs, with each from one domain. Then, the framework is constructed with three main components: 1) A bidirectional image-to-image translator is firstly used to generate two synthesized images from the two input images. Then, 2) four shared-weighted segmentation networks are performed following the translator to segment over all the four images, including all the input images and the synthesized images. Finally, 3) A global-to-local adaptor is inserted between the segmentation networks to reduce the domain shift.

\subsection{Bidirectional Cross-modal Image Synthesis}

Domain shift is large between the source ($T_1$) and target domain ($T_2$) for the brain tumor segmentation task in multi-modal MR images. Several studies \cite{cycada,SIFA} proposed to use image translator to firstly reduce this gap. Therefore, we also tackle this domain shift problem by transferring image appearances to the counterpart domains before align the distributions of the two domains. Specifically, given the unpaired inputs $x_s \in S$, $x_t \in T$ for the source and target domains, two synthesized images $x_{s\to t}$ and $x_{t\to s}$ are produced using an unsupervised image synthesis network proposed in \cite{zhou2020hi}, with an pix2pix \cite{zhu2017unpaired} like generative adversarial learning paradigm. Specifically, for images from $x_s$, the generator $G$ of this method tries to generate images $x_{s \to t}$ similar to $x_t$. The objective function of the generator can be therefore written as,
\begin{align}
    \mathcal{L}^G = \mathbb{E}_{x_s \in S}{\log (1 - D(x_s, x_{s \to t})} + \lambda \mathbb{E}{\||x_t - x_{s \to t}\||},
\end{align}
\noindent where $x_{s \to t} = G(x_s)$. And the discriminator $D$ tries to discriminate these two images, and can be written as,
\begin{align}
    \mathcal{L}^D = \mathbb{E}{\log D(x_t)} - \mathbb{E}{\log (1 - D(x_{s \to t})}.
\end{align}

Obviously, the synthesized images convey the style information of the new modality and the texture information of the old one. Therefore, they can be regarded as intermediate samples for transferring from the source to the target domain. Afterward, different from the aforementioned methods that using only the synthesized target images (\ie, \cite{cycada,SIFA-TMI}) for domain alignment, our method takes the effect of the source images to provide structural consistencies also in the source domain. In this case, our method also establishing a bidirectional adaptation scheme, other than the bidirectional image synthesis scheme. Then, the four images (\ie, two real and two synthesized images) are fed into an attention-based encoder-decoder network for segmentation.

\subsection{Dual-attention based Segmentation Network}

U-Net based models have been widely applied in brain tumor segmentation. However, it cannot identify the discriminative features for capturing high-level semantic cues. To suppress irrelevant regions while highlighting useful features for medical segmentation task, attention mechanisms (including Squeeze-and-Excitation mechanisms \cite{chen2017sca}) have been widely used in convolutional neural networks to enhance such an ability. Besides, the attention U-Net \cite{schlemper2019attention} is proposed to incorporate a gated attention mechanism into each level of the conventional U-Net in the up-sampling path. 
By contrast, we equip our U-Net with the dual-attention mechanism proposed in \cite{fu2019dual} which applies not only spatial-wise but also channel-wise attention for high-level features. The dual-attention block is inserted in-between the encoding and decoding path of U-Net. The dual-attention U-Net can therefore effectively learn the feature relationships of high-level extracted features. Specifically, given a feature of $C$ channels with spatial dimensions of $H\times W$. The position-wise attention mechanism learns an $H\times W\times H\times W$ dimensional mask to reveal the objective-oriented feature relationships inside each feature map. Similarly, the channel-wise attention module of $C\times C$ dimensions reveals the salience in-between the feature maps. Given $\xi$ as the input of the attention block, the output of each attention masked feature $\xi^A$ can be obtained by
\begin{align}
    \xi^A_{Pos.} = \alpha \sum_{k=1}^{H\times W} (A \xi^B) + \xi, \xi^A_{Cha.} = \alpha \sum_{k=1}^{C} (A \xi^B) + \xi,
\end{align}
\noindent where $A$ denote the abovementioned attention masks for each module, $\xi^B$ denote the feature before the attention block, and $\alpha$ is a learnable weight. The output of the dual-attention module is the aggregation of the two attention masked features. More importantly, our experiments show that the proposed dual-attention U-Net improves the segmentation performance compared to the conventional U-Net model (See Section \ref{Sec:Exp} for details).

\textbf{Remarks}. The motivation behind the attention-based U-Net is based on the supposition that the attention mask encodes the most discriminative image semantics, which is helpful for achieving an accurate adaptation between the two domains. Since the two domains cannot be fully-aligned with each other, for which existing UDA methods \cite{cycada,output,SIFA} tend to incorporate significant noise and inconsistency when using features or output masks for the alignment. Experiments show that our proposed attention-based alignment led to an more effective adaptation scheme (See comparisons in Section \ref{Sec:Ab}).

\subsection{Global-to-local Adaptation Module}

With the cross-modality image synthesis, the distribution of the two domains are more likely to be located close together in the feature space. In addition, we propose a group of global-to-local alignment constraints to align the features of the images. These constraints include the output-space, feature-level, and attention-level consistency learning (denote as $\mathcal{L}^{Att.}_{consis}$) which is applied on the masks (denote as $\mathbf{A}^{Att.}$) learned by the two kinds of attention masks (denote as $Att.= \{Position,Channel\}$). Following the existing domain alignment methods \cite{SIFA-TMI}, we perform adversarial learning for the alignment of features and attention masks.

Different from the purely image-to-image translation task, the cross-domain segmentation task requires the real and fake images to maintain consistent semantic information. This constraints it to produce the same segmentation results. Therefore, for $x_s$ and $x_{s \to t}$, we first calculate the loss $\mathcal{L}_{seg}^s$ and ${L}_{seg}^{Syn.s}$ using the cross-entropy loss and the generalized dice loss with a class-balance weight of $w^i=1/num(x^i)$. Here, $i \in \{1,...,c\}$ denote the $i$th of $c$ classes. Thus, the two losses can be formulated as follows:
\begin{equation}
\small
\left\{
\begin{aligned}
 &\mathcal{L}_{seg}^s = -\mathbb{E}_{\mathbf x_s, \mathbf y_s\sim S}\Bigl[\mathbf y_s \log \Phi\left(\mathbf x_s \right)  + 1 - \sum_{i=1}^c{2w^i\frac{\Phi \left(\mathbf x_s^i\right)\mathbf y^i_s}{\Phi \left(\mathbf x^i_s\right) + \mathbf y^i_s}} \Bigr], \\
&\mathcal{L}_{seg}^{Syn.s} = -\mathbb{E}_{\mathbf x_{s\to t}, \mathbf y_s\sim T'}\Bigl[\mathbf y_s \log \Phi\left(\mathbf x_{s\to t}  \right) + 1 - \sum_{i=1}^c{2w^i\frac{\Phi \left(\mathbf x_{s\to t}^i\right)\mathbf y^i_s}{\Phi \left(\mathbf x^i_{s\to t}\right) + \mathbf y^i_s}} \Bigr]. 
\end{aligned}
\right.
\end{equation}

Then, the consistency for $x_s$ and $x_{s\to t}$ is achieved by giving them the same ground-truths. The consistency for $x_t$ and $x_{t \to s}$ is achieved by applying the output consistency constraint $\mathcal{L}^{output}_{consis}$, since they do not have the ground-truth masks in training. This loss can be formulated by 
\begin{align}
  \mathcal{L}^{output}_{consis} = -\mathbb{E}_{\mathbf x_t\sim T, \mathbf x_{t\to s}\sim S'}\Bigl[\left\| \Phi\left(\mathbf x_t  \right) - \Phi \left(\mathbf x_{t\to s} \right) \right\|_2\Bigr],
\end{align}
\noindent where $y_s$ denote the ground-truth label for the source image, and $\Phi\left(\mathbf \cdot  \right)$ denote the overall operation of the abovementioned feature extractor.

In the meanwhile, the attention masks (\ie, $\mathbf{m}^{Att.}_t$ and $\mathbf{m}^{Att.}_{t\to s}$) of unlabeled images are forced to be closer to the labeled images (\ie, $\mathbf{m}^{Att.}_s$ and $\mathbf{m}^{Att.}_{s\to t}$). We achieved this domain alignment using adversarial learning. In this learning paradigm, the segmentation network is regarded as the generator to generate segmentation mask for the images from different images in a certain domain, and the discriminator $D$ tries to discriminate the attention masks of the different input images. Therefore, the consistency $\mathcal{L}^{Att.}_{consis}$ for the source and target domain can be defined as follows:
\begin{equation}
\begin{aligned}
\mathcal{L}^{Att.}_{consis} = &\mathbb{E}\Bigl[\sum \mathbf y^{Att.}_s \log\left(D_s^{Att.}\left(\mathbf{A}^{Att.}_{t\to s}\right) \right)\Bigr]  \\ & + \mathbb{E}\Bigl[\sum \mathbf y^{Att.}_{s\to t} \log\left(D_t^{Att.}\left(\mathbf{A}^{Att.}_t \right)\right) \Bigr],
\end{aligned}
\end{equation}
\noindent where $\mathbf{A}^{Att.}$ denote the attention masks, where $D_*^{Att.}$ denote the feature of discriminators in domain $* \in \{source,target\}$. Besides, the adversarial loss for the source (\ie, $\mathcal{L}s^{Att.}_{adv}$) and target (\ie, $\mathcal{L}t^{Att.}_{adv}$) can be defined by
\begin{equation}
\small
\left\{
\begin{aligned}
& \mathcal{L}s^{Att.}_{adv} = \mathbb{E}\Bigl[\log\left(D_s^{Att.}\left(\mathbf{A}^{Att.}_s \right) \right)\Bigr]  + \mathbb{E}\Bigl[\log\left(1 - D_s^{Att.}\left(\mathbf{A}^{Att.}_{t\to s} \right) \right) \Bigr],\\
& \mathcal{L}t^{Att.}_{adv} = \mathbb{E}\Bigl[\log\left(D_t^{Att.}\left(\mathbf{A}^{Att.}_{s\to t} \right) \right)\Bigr]  + \mathbb{E}\Bigl[\log\left(1 - D_t^{Att.}\left(\mathbf{A}^{Att.}_t \right) \right) \Bigr],
\end{aligned}
\right.
\end{equation}
\noindent where $y_*$ is the indicator, where $y_* = 1$ if the mask comes from the same domain. To leverage more information between domains, we also align the features after the attention block for the intra-domain images. This is commonly used as feature-level alignment in other UDA methods, eg, \cite{cycada}. Similarly, the constraints of consistency loss and adversarial loss on the two domains for features can be defined by
\begin{equation}
\begin{aligned}
\mathcal{L}^{Feat.}_{consis} = &\mathbb{E}\Bigl[\sum \mathbf y^{Feat.}_s \log\left(D_s^{feat.}(\mathbf{\xi}_{t\to s}) \right)\Bigr] \\ &+\mathbb{E}\Bigl[\sum \mathbf y^{Feat.}_{s\to t} \log\left(D_t^{feat.}(\mathbf{\xi}_t) \right) \Bigr],
\end{aligned}
\end{equation}
\begin{equation}
\left\{
\begin{aligned}
 & \mathcal{L}s^{Feat.}_{adv} = \mathbb{E}\Bigl[\log\left(D_s^{Feat.}\left(\mathbf{\xi}_s \right) \right)\Bigr] + \mathbb{E}\Bigl[\log\left(1 - D_s^{Feat.}\left(\mathbf{\xi}_{t\to s} \right) \right) \Bigr],\\
 & \mathcal{L}t^{Feat.}_{adv} = \mathbb{E}\Bigl[\log\left(D_t^{Feat.}\left(\mathbf{\xi}_{s\to t} \right) \right)\Bigr] + \mathbb{E}\Bigl[\log\left(1 - D_t^{Feat.}\left(\mathbf{\xi}_t \right) \right) \Bigr].
\end{aligned}
\right.
\end{equation}

\subsection{Loss Function}

Finally, we formulate the tumor segmentation, consistency learning and adversarial learning into a unified framework based on a UDA strategy. Thus, we can obtain the total loss by a weighted summation of all the previously defined loss functions, which is given by 
\begin{equation}
\begin{aligned}
  \mathcal{L}_{total} = \mathcal{L}_{seg} + &\lambda_{out} \mathcal{L}^{output}_{consis} + \lambda_{syn} (\mathcal{L}^{G} + \mathcal{L}^{D})  \\ + &\lambda_{gtl} (\mathcal{L}^{Feat.}_{adv} +  \mathcal{L}^{Feat.}_{consis} + \mathcal{L}^{Att.}_{adv} + \mathcal{L}^{Att.}_{consis}),
\end{aligned}
\end{equation}

\noindent where $\mathcal{L}^{G}$ and $\mathcal{L}^{D}$ are the losses defined in the image synthesis network. The weights of these losses are set to $0.001,0.1,0.01$ for $\lambda_{out}, \lambda_{gtl}$, and $\lambda_{syn}$, respectively.

\subsection{Implementation Details}

Our method is implemented using the open-source framework \emph{PyTorch}. 
The experiments are accelerated by four Nvidia Tesla V100 GPUs, with $32$ GB memory for each. 
Note that the proposed framework can be trained end-to-end. However, it is not efficient to train the image synthesis module and domain adaptation module simultaneously, as limited by the GPU memory and time. Therefore, in our experiments, we first train the image synthesis module, and then fix it to train the segmentation and domain adaptation module. 
Unlike the method in \cite{cycada} which trains the generator after the discriminator becomes too strong, we train the backbone and the discriminators sequentially in each iteration. 
As the images are distinct from natural images, we do not use the pre-trained models from outside large scale datasets, \eg, ILSVRC, and randomly initialize the network parameters. 
We use the learning rate of $5\times10^{-3}$ to train our network, which is decayed by a polynomial policy with a power of $0.75$. We train the model over $150$ epochs. The learning rate of the discriminator is set to $5e-5$ without learning rate decay. The weights for segmentation consistency, feature consistency, position- and channel-wise attention consistency are set to $0.001, 0.1, 0.1, 0.1$, respectively. 

\begin{table*}[htbp]
\renewcommand{\arraystretch}{1.0}
\centering
\caption{\label{Table:sotaDSC} Comparison our model with the state-of-the-art methods in DSC (\%) for the task of $T1$ to $T2$. The best results are highlighted in \textbf{bold}.} \vspace{0.15cm}
\setlength{\tabcolsep}{8pt}
\begin{tabular}{c|ccc|c}
\toprule[1pt]
\textbf{Method} & WT & TC & ET & Average \\
\toprule[1pt]
\textbf{Source Only} &14.95$\pm$15.10 &14.71$\pm$16.45 &2.91$\pm$6.97 & 10.86\\
\hline
\textbf{AdaptSegNet \cite{output}} &65.70$\pm$12.35 &52.51$\pm$22.86 &27.79$\pm$19.37 & 48.67\\
\textbf{Cycada \cite{cycada}} &67.87$\pm$18.33 &57.33$\pm$25.63 &33.99$\pm$21.39 & 53.06\\
\textbf{SIFA-TMI \cite{SIFA-TMI}} & 70.81$\pm$19.86  & 65.32$\pm$24.57  & 44.46$\pm$23.88  & 60.20 \\
\hline
\textbf{BiGL (Ours)} & \textbf{80.39$\pm$12.48} & \textbf{74.76$\pm$22.21} & \textbf{53.73$\pm$25.97} & \textbf{69.63} \\
\hline \hline
\textbf{Target Only (Multi-Modal)} & 88.60$\pm$7.31  & 83.74$\pm$13.45  & 75.96$\pm$24.02  & 82.77 \\
\toprule[1pt]
\end{tabular}\\
\end{table*}
\begin{table*}[htbp]
\renewcommand{\arraystretch}{1.0}
\centering
\caption{\label{Table:sotaHD95} Performance comparison with the state-of-the-art methods in HD95 (mm) for the task of $T1$ to $T2$. The best results are highlighted in \textbf{bold}.} \vspace{0.15cm}
\setlength{\tabcolsep}{8pt}
\begin{tabular}{c|ccc|c}
\toprule[1pt]
\textbf{Method} & WT & TC & ET & Average \\
\toprule[1pt]
\textbf{Source Only} &55.89$\pm$17.29&68.78$\pm$18.19 &61.56$\pm$26.49 & 62.08\\
\hline
\textbf{AdaptSegNet (\cite{output})} &25.88$\pm$19.15 &29.45$\pm$21.56 &29.45$\pm$21.07& 28.26 \\
\textbf{Cycada (\cite{cycada})} &31.23$\pm$21.70 &28.94$\pm$23.96 &27.71$\pm$25.47& 29.29 \\
\textbf{SIFA-TMI (\cite{SIFA-TMI})} & 14.89$\pm$ 15.77 &19.29 $\pm$17.43  & \textbf{17.46$\pm$ 17.72} & 17.21 \\
\hline
\textbf{BiGL (Ours)} &\textbf{13.19$\pm$13.74} &\textbf{17.38$\pm$17.47} &{18.23$\pm$19.23} & \textbf{16.27} \\
\hline \hline
\textbf{Target Only (Multi-Modal)} & 11.49$\pm$17.05  & 11.19$\pm$12.59  & 13.62$\pm$21.79  & 12.10 \\
\toprule[1pt]
\end{tabular}\\
\end{table*}

\section{Experimental Results}
\label{Sec:Exp}

\subsection{Datasets}
We use the multi-modal Brain Tumor Segmentation Challenge 2019 (BraTS’ 19) \cite{bakas2018identifying} dataset to evaluate the effectiveness of the proposed model. This dataset consists of 335 patients with manual delineations of three lesion regions, \ie, the necrotic and non-enhancing tumor core (NCR/NET), peritumoral edema (ED) and GD-enhancing tumor (ET). Each patient has multiple MR images, including $T_1$, $T_2$, $T_1$Gd and $T_2$ FLAIR with the size of $240\times 240\times 155$. These images are acquired from $19$ institutions. We perform the experiments on adapting the segmentation network from $T_1$ to $T_2$ as our task, \ie, the network is trained on $T_1$ images and tested on $T_2$ images. We randomly partition the dataset by 70$\%$,10$\%$ and 20$\%$ for training, validation and testing, respectively. The challenge evaluates the segmentation performance on three regions, \ie, whole tumor (WT), tumor core (TC), and enhancing tumor (ET). In particular, WT includes the areas of NCR/NET, ED and ET. TC includes the area of NCR/NET and ET. We follow this protocol and report the results in terms of Dice Similarity Coefficient (DSC) and 95$\%$ Hausdorff Distance (HD95), as also required by the challenge.

\subsection{Comparison with State-of-the-Art Methods}

\begin{figure*}[!t]
  \centering
  \includegraphics[width=\linewidth]{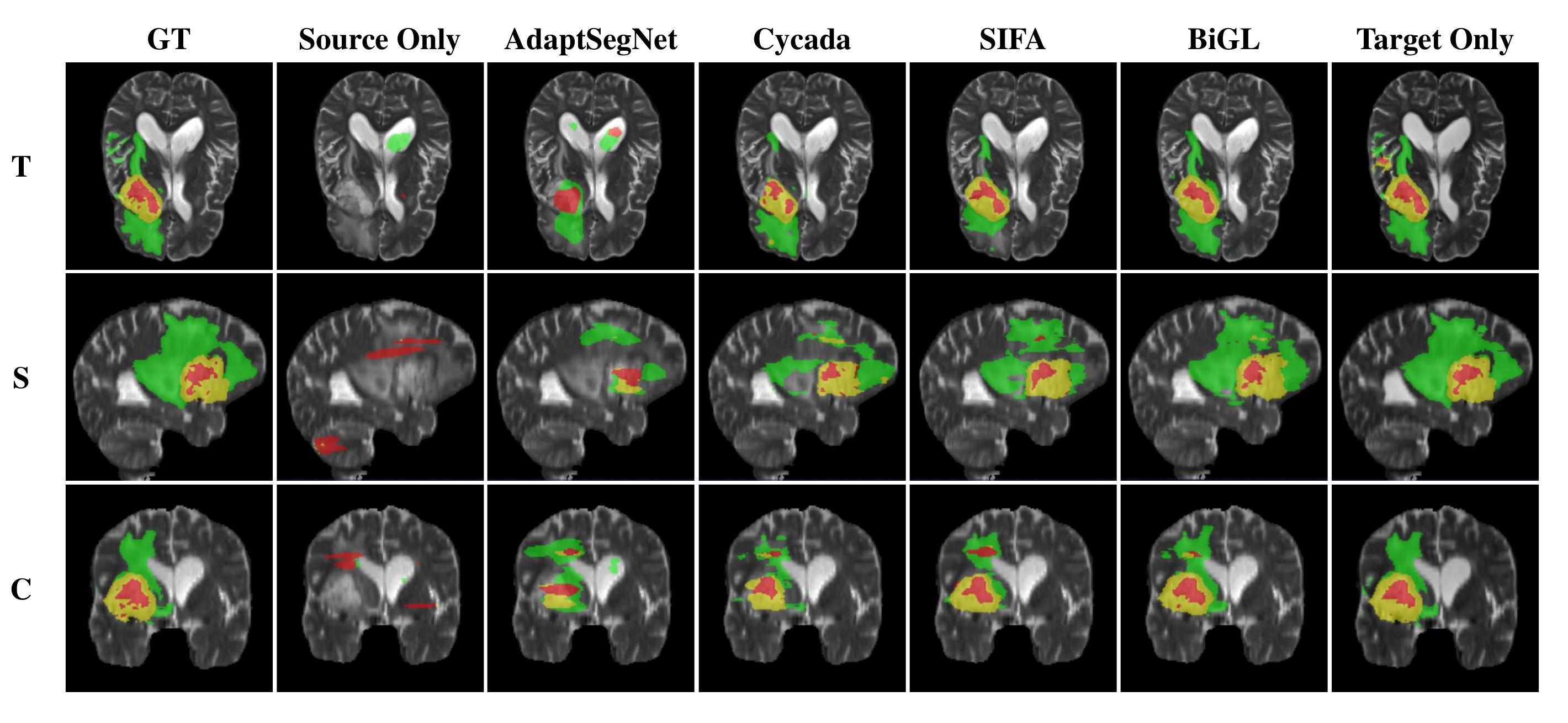} \vspace{-0.75cm}
    \caption{\label{fig:vis} The visualization results of competing methods for a typical case in three view angles. Green denote WT, yellow denote TC, and red denote ET. 'T' denote the transverse plane, 'S' denote the saggital plane, 'C' denote the coronal plane.}
\end{figure*}

We compare against the state-of-the-art unsupervised domain adaptation methods for segmentation. The methods are designed not only for medical images \cite{SIFA}, but also for natural images \cite{cycada,output}. Specifically, the method in \cite{cycada} (denoted as Cycada) leverages an unsupervised image-to-image translator to generate a synthesized distribution that is closer to the target domain. The method in \cite{output} (denoted as AdaptSegNet) assumes that the structured outputs of the two domains are similar. Finally, the method in \cite{SIFA-TMI} (denoted as SIFA-TMI) combines the strategies of the abovementioned two methods. To determine the lower and upper bound of the UDA methods, we further conduct 1) the "Source Only" method, which is an unadapted method that trains the network on $T_1$ images and directly applies it to $T_2$ images, and 2) the "Target Only" method, which is fully-supervised on all the image modalities provided by the dataset (\ie, T1, T1c, T2 and FLAIR). 

The performance comparison in terms of DSC and HD95 is reported in Table \ref{Table:sotaDSC} and \ref{Table:sotaHD95}. As can be seen, the domain gap between the $T_1$ and $T_2$ images in the brain tumor segmentation task is huge, with the Source Only method only obtaining an average performance of $10.86\%$ in terms of DSC and $62.08$ in terms of HD95. Compared with Source Only method, all UDA methods (in the second set of rows) have significantly improved the performance, with over $30\%$ increase in DSC. And compared with AdaptSegNet, the performance of Cycada boosted the performance by $2.17\%$, $4.82\%$, and $6.20\%$ for WT, TC and ET, respectively, with $4.39\%$ in average. This suggest the effectiveness of the proposed image synthesis module for firstly reducing the domain gap by transferring the image appearances. 
Among the previous methods, SIFA network performs best with $60.20\%$ and $17.21$ in DSC and HD95, respectively. Compared with this method, BiGL network improves the performance greatly, increasing $9.43\%$ in terms of DSC and decreasing $0.94$ in terms of HD95. Besides, the performance of BiGL in terms of DSC consistently outperforms the previously best performing method SIFA in all the three classes. For the performance in HD95, BiGL also outperforms SIFA by decreasing the value from $14.89$ and $19.29$ to $13.19$ and $17.38$ for WT and TC, respectively. And it also performing comparable results for ET with $18.23$, and finally get a $0.94$ decreasing in average.

\subsection{Visualization Results}

We compare the segmentation results by BiGL with those produced by the state-of-the-art methods in Fig. \ref{fig:vis}. As can be seen, the Source Only method cannot generate segmentation as influenced by the huge domain gap. The AdaptSegNet tend to over-segment on TC, while generating less segmentations for WT. The SIFA (\ie, SIFA-TMI) method performs better in this case. However, it still generate less segmentations in both categories. The proposed method generates more precise results through all three classes.

\subsection{Ablation Study}
\label{Sec:Ab}

\begin{table*}[htbp]
\setlength{\tabcolsep}{8pt}
\renewcommand{\arraystretch}{1.0}
\centering
\caption{\label{Table:Effect} Evaluating the effectiveness of the proposed BiGL method in terms of \textbf{DSC} (\%), w/ or w/o output, feature and attention consistencies.} \vspace{0.15cm}
\begin{tabular}{ccc|ccc|c}
\toprule[1pt]
output & feature & attention & WT & TC & ET & Average \\
\toprule[1pt]
\checkmark &&& 73.26$\pm$18.87 & 63.91$\pm$24.38 &41.34$\pm$23.13 & 59.50 \\
\checkmark&\checkmark&& 76.72$\pm$14.40 & 70.56$\pm$21.83 &48.67$\pm$24.38 & 65.32 \\
\checkmark &\checkmark&\checkmark & \textbf{80.39$\pm$12.48} & \textbf{74.76$\pm$22.21} & \textbf{53.73$\pm$25.97} & \textbf{69.63} \\
\toprule[1pt]
\end{tabular}
\end{table*}

\begin{table*}[htbp]
\setlength{\tabcolsep}{8pt}
\renewcommand{\arraystretch}{1.0}
\centering
\caption{\label{Table:EffectA} Evaluating the effectiveness of the proposed BiGL method in terms of \textbf{HD95} (mm), w/ or w/o output, feature and attention consistencies.}
\begin{tabular}{ccc|ccc|c}
\toprule[1pt]
output & feature & attention & WT & TC & ET & Average\\
\toprule[1pt]
\checkmark &&& 46.09$\pm$40.65 &66.80$\pm$50.28 &22.47$\pm$22.54 & 45.12 \\
\checkmark&\checkmark&&14.11$\pm$13.88 &19.69$\pm$17.87 &\textbf{17.33$\pm$16.62} & 17.04 \\
\checkmark &\checkmark&\checkmark &\textbf{13.19$\pm$13.74} &\textbf{17.38$\pm$17.47} &18.23$\pm$19.23 & \textbf{16.27} \\
\toprule[1pt]
\end{tabular}
\end{table*}

In the ablation study, we first evaluate the effectiveness of the proposed consistency learning constraints. The comparison of segmentation results in terms of DSC and HD95 are shown in Table \ref{Table:Effect} and \ref{Table:EffectA}, respectively. In the different rows, the proposed method is equipped with or without output, feature and attention consistencies. 
From these results, it can be observed that our full model with the output, feature and attention consistencies performs better than all competing methods. Besides, when incorporating more consistency learning constraints (see the third vs. second row in Table \ref{Table:Effect}), the performance of the proposed method is consistently improved. By adding the attention consistency, the performance increases from $65.32\%$ to $69.63\%$ in terms of DSC, and decreases from $17.04$mm to $16.27$mm in terms of HD95, which validates the effectiveness of the proposed global-to-local adaptation strategy.

We further evaluate the influence of the weight $\lambda_{gtl}$ with selected numbers $\{0.02,0.1,0.5\}$. The results are reported in Table \ref{Table:lambda}. As suggested by the table, the model achieves the best overall performance with a weight of $0.1$. Another observation is that, the DSC is not sensitive to the weight $\lambda_{gtl}$. By contrast, as an indicator of the unique values in segmentation, HD95 is sensitive to the weight for feature and attention constraints. This reveals the effectiveness of these alignments for local representations.

\begin{table*}[htbp]
\setlength{\tabcolsep}{10pt}
\renewcommand{\arraystretch}{1.0}
\centering
\caption{\label{Table:lambda} The influence of the weight $\lambda_{gtl}$.}
\begin{tabular}{clccccccc}
\toprule[1pt]
\multirow{2}{*}{Method} & \multirow{2}{*}{Param.} & \multicolumn{3}{c}{DSC} && \multicolumn{3}{c}{HD95} \\
\cline{3-5}
\cline{7-9}
&& WT & TC & ET && WT & TC & ET \\
\toprule[1pt]
\textbf{BiGL} & $\lambda_{gtl} = 0.02$ & 78.38 & 71.52 & 50.14 &&17.26&19.87& \textbf{17.47}\\
& $\lambda_{gtl} = 0.1$ & \textbf{80.39} & \textbf{74.76} &\textbf{53.73}&&\textbf{13.19}&\textbf{17.38}&18.23\\
& $\lambda_{gtl} = 0.5$ & 79.62 & 73.17 &52.53&&25.13&22.23&17.81\\
\hline
\toprule[1pt]
\end{tabular}
\end{table*}

\begin{figure*}[!t]
  \centering
  \includegraphics[width=\linewidth]{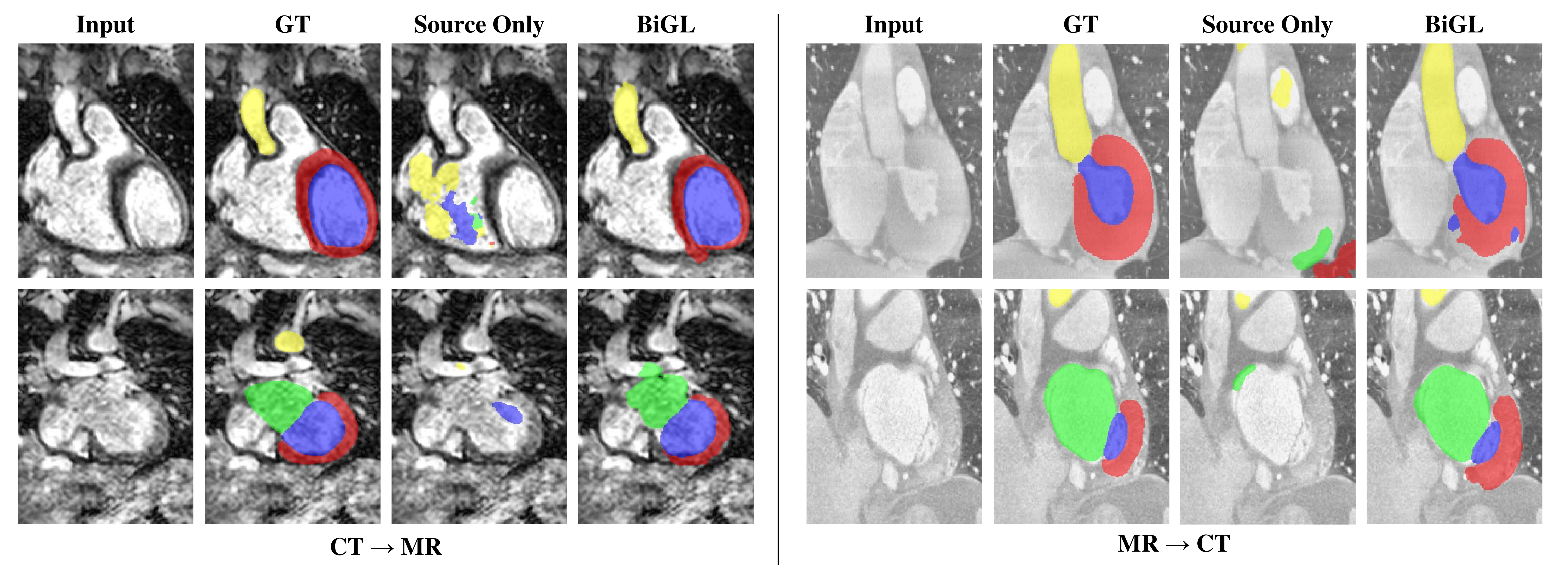}
    \caption{\label{fig:viscard} Visualization results for two typical cases in the MMWHS dataset.}
\end{figure*}

\begin{table*}[htbp]
\renewcommand{\arraystretch}{1.0}
\centering
\caption{\label{Table:cardDSC} Performance comparison with the state-of-the-art methods in terms of \textbf{DSC} for the task of cardiac segmentation. (The best results are highlighted in \textbf{bold}.)}
\setlength{\tabcolsep}{5pt}
\begin{tabular}{c|cccc|c|cccc|c}
\toprule[1pt]
\textbf{Task} & \multicolumn{5}{c}{CT to MR} & \multicolumn{5}{c}{MR to CT} \\
\hline
\textbf{Method} & AA & LAC & LVC & MYO & Avg. & AA & LAC & LVC & MYO & Avg.\\
\toprule[1pt]
\textbf{Source Only(SIFA-TMI)} & 5.4 & 30.2 & 24.6 & 2.7 & 15.7 & 28.4 & 27.7 & 4.0 & 8.7 & 17.2 \\
\textbf{Source Only} & 14.5 & 14.4 & 26.6 & 1.3 & 14.2 & 38.9 & 57.8 & 5.2 & 10.8 & 28.2\\
\hline
\textbf{AdaptSegNet (\cite{output})} & 60.8 & 39.8 & 71.5 & 35.5 & 51.9 & 65.2 & 76.6 & 54.4 & 43.6 & 59.9 \\
\textbf{CycleGAN (\cite{zhu2017unpaired})} & 64.3 & 30.7 & 65.0 & 43.0 & 50.7 & 73.8 & 75.7 & 52.3 & 28.7 & 57.6 \\
\textbf{Cycada (\cite{cycada})} & 60.5 & 44.0 & 77.6 & \textbf{47.9} & 57.5 & 72.9 & 77.0 & 62.4 & 45.3 & 64.4 \\
\textbf{SIFA-AAAI (\cite{SIFA})} & \textbf{67.0} & 60.7 & 75.1 & 45.8 & 62.1 & 81.1 & 76.4 & 75.7 & 58.7 & 73.0 \\
\textbf{SIFA-TMI (\cite{SIFA-TMI})} & 65.3 & 62.3 & 78.9 & 47.3 & 63.4 & 81.3 & \textbf{79.5} & 73.8 & \textbf{61.6} & 74.1 \\
\hline
\textbf{BiGL (Ours)} & 62.0 & \textbf{71.1} & \textbf{88.0} & 43.7 & \textbf{66.2} & \textbf{87.7} & 78.6 & \textbf{80.0} & 54.9 & \textbf{75.3} \\
\hline
\textbf{Target Only(SIFA-TMI)} & 82.8  & 80.5  & 92.4 & 78.8 & 83.6 & 92.7 & 91.1 & 91.9 & 87.7 & 90.9 \\
\toprule[1pt]
\end{tabular}\\
\end{table*}

\begin{table*}[htbp]
\renewcommand{\arraystretch}{1.0}
\centering
\caption{\label{Table:cardASD} Performance comparison with the state-of-the-art methods in terms of \textbf{ASD} for the task of cardiac segmentation. (The best results are highlighted in \textbf{bold}.)} \vspace{0.15cm}
\setlength{\tabcolsep}{5pt}
\begin{tabular}{c|cccc|c|cccc|c}
\toprule[1pt]
\textbf{Task} & \multicolumn{5}{c}{CT to MR} & \multicolumn{5}{c}{MR to CT} \\
\hline
\textbf{Method} & AA & LAC & LVC & MYO & Avg. & AA & LAC & LVC & MYO & Avg. \\
\toprule[1pt]
\textbf{Source Only(SIFA-TMI)} & 15.4 & 16.8 & 13.0 & 10.8 & 14.0 & 20.6 & 16.2 & N/A & 48.4 & N/A \\
\textbf{Source Only} & 22.4 & 27.6 & 10.3 & 10.7 & 17.8 & 26.8 & 13.3 & 17.7 & 20.2 & 19.5 \\
\hline
\textbf{AdaptSegNet (\cite{output})} & \textbf{5.7} & 8.0 & 4.6 & 4.6 & 5.7& 17.9 & 5.5 & 5.9 & 8.9 & 9.6 \\
\textbf{CycleGAN (\cite{zhu2017unpaired})} & 5.8 & 9.8 & 6.0 & 5.0 & 6.6 & 11.5 & 13.6 & 9.2 & 8.8 & 10.8 \\
\textbf{Cycada (\cite{cycada})} & 7.7 & 13.9 & 4.8 & 5.2 & 7.9 & 9.6 & 8.0 & 9.6 & 10.5 & 9.4 \\
\textbf{SIFA-AAAI (\cite{SIFA})} & 6.2 & 9.8 & 4.4 & 4.4 & 6.2 & 10.6 & 7.4 & 6.7 & 7.8 & 8.1\\
\textbf{SIFA-TMI (\cite{SIFA-TMI})} & 7.3 & 7.4 & 3.8 & 4.4 & 5.7 & \textbf{7.9} & 6.2 & \textbf{5.5} & 8.5 & 7.0 \\
\hline
\textbf{BiGL (Ours)} & 10.9 & \textbf{5.4} & \textbf{2.3} & \textbf{3.3} & \textbf{5.5} & 8.0 & \textbf{5.0} & 6.1 & \textbf{5.6} & \textbf{6.2} \\
\hline
\textbf{Target Only(SIFA-TMI)} & 3.6  & 3.9  & 2.1 & 1.9 & 2.9 & 1.5 & 3.5 & 1.7 & 2.1 & 2.2 \\
\toprule[1pt]
\end{tabular}\\
\end{table*}

\subsection{Evaluation on Cardiac Substructure Segmentation Dataset}

The generalization ability of the UDA methods is obviously more important than the fully-supervised methods. Therefore, we employed the Multi-Modality Whole Heart Segmentation (MMWHS) Challenge 2017 dataset for cardiac substructure segmentation as the benchmark dataset for additional evaluation. We use the training dataset which contains $20$ MR and $20$ CT images, for a fair comparison to SIFA-AAAI. In this dataset, besides the previously mentioned methods, we also involved the method in \cite{zhu2017unpaired} (denoted as CycleGAN) and \cite{SIFA} (denoted as SIFA-AAAI). CycleGAN uses purely the image translator for domain adaptation. The target images are transferred to the source modality, and then segmented by the network trained in the source domain. SIFA-AAAI is the conference version of the method SIFA-TMI, with the comparable results to its journal version. For fair comparisons, we use the same training and testing data split to SIFA.

The segmentation performance for UDA task of \{CT to MR\} and \{MR to CT\} in terms of DSC and ASD are reported in Table \ref{Table:cardDSC} and \ref{Table:cardASD}, respectively. As suggested by the tables, Source Only performs also worse with a non-adaptation fashion, but is a little better than that in the case of brain tumor segmentation. This indicates the challenges of brain tumor segmentation task. For the task of CT to MR UDA segmentation task, AdaptSegNet and CycleGAN both performs better than Source Only, validating the effectiveness of image translator and output space domain alignment. Combining these two techniques, Cycada and SIFA-AAAI gain significant improvements of $5.6\%$ and $10.2\%$ in DSC, compared with AdaptSegNet, respectively. By incorporating the output space alignment in multiple scale, SIFA-TMI still improves the performance by $1.3\%$ in DSC. However, as these three methods (\ie, Cycada, SIFA-AAAI and SIFA-TMI) do not consider the local consistencies in domain alignment, they do not perform better than AdaptSegNet in terms of ASD. Compared with these methods, the proposed BiGL method performs consistently best in both DSC and ASD, showing the effectiveness of the proposed bidirectional global to local domain alignment. For the task of MR to CT, BiGL has achieved significant performance improvement on DSC and ASD, with $75.3\%$ and $6.2$, respectively. Moreover, the performance of BiGL on specific classes of AA and LVC is significantly better than previous methods. BiGL outperforms SIFA-TMI by $6.4\%$ for AA in terms of DSC, from $81.3\%$ to $87.7\%$, and $6.2\%$ for LVC, from $73.8\%$ to $80.0\%$.

To visualize the performance of BiGL, we also show the segmentation results for two typical cases, as displayed in Fig. \ref{fig:viscard}. According to the figure, Source Only method still generate worse results, with almost wrong predictions in all four classes. Although in an unsupervised setting, BiGL method also performs well through these cases, and generates more complete contours and organ structures with high overlapping to the ground-truth.

\section{Conclusion}
\label{sec:conclusion}

In this paper, we have proposed an unsupervised attention domain adaptation method for cross-modality brain tumor segmentation. Our method integrates both bidirectional image-to-image transfer and representation alignments from global-to-local perspective, which can obtain better model generalization under the situation of large domain gaps. Specifically, our method leverages synthesized images for both modalities to produce intermediate data distributions between the source and target domains. 
In addition, we propose an attention adaptation to achieve an efficient adaptation. 
The experimental results demonstrate the effectiveness of the proposed model over the other state-of-the-art methods.
Moreover, our framework is quite flexible and general, and can easily be extended to different UDA tasks for other modalities, such as $T_1$ and $Flair$, etc. 
Although the proposed method has obtained better performance than other methods, there is still significant room for improvement. Therefore, more precise domain alignment algorithms will be developed in future work.


%

\ifCLASSOPTIONcompsoc
  \section*{Acknowledgments}
\else
  \section*{Acknowledgment}
\fi

The authors would like to thank...

\ifCLASSOPTIONcaptionsoff
  \newpage
\fi



\bibliographystyle{IEEEtran}
\bibliography{IEEEabrv,UDASEG}
\end{document}